\documentclass[12pt]{article}
\usepackage[utf8]{inputenc}
\usepackage[T1]{fontenc}
\usepackage{theorem}
\usepackage{bbm}
\usepackage{mathtools}
\usepackage{multirow}
\usepackage{amssymb}
\usepackage{array}
\usepackage{color}
\usepackage{graphicx}
\usepackage{pgfplots}
\usepackage{tikz}
\usetikzlibrary{positioning}
\usepackage[export]{adjustbox}
\usetikzlibrary{shapes.geometric, arrows}
\usepackage{amsmath}
\usepackage{verbatim} 
\usepackage{rotating}
\usepackage[round]{natbib}
\usepackage[colorlinks=true,linkcolor=black, citecolor=blue, urlcolor=blue]{hyperref}
\usepackage{footnote}
\usepackage{tcolorbox}
\usepackage{placeins}
\makesavenoteenv{tabular}
\makesavenoteenv{table}

\newtheorem{theorem}{Theorem}

\newtheorem{remark}{Remark}
\newtheorem{prop}{Proposition}

\newcommand*{\QED}{\hfill\ensuremath{\blacksquare}}
\DeclareMathOperator*{\argmin}{argmin} 

\newtheorem{model}{\textbf{Model}}

\graphicspath{D:/PhD-Works/Articles/Thirds/Article3/figures}

\begin{document}
\begin{center}
	\large{{\scshape Consensual Aggregation on Random Projected High-dimensional Features for Regression}}\\
	\bigskip
	\normalsize
	Sothea Has
\end{center}

\begin{flushleft}
LPSM, Sorbonne Université Pierre et Marie Curie (Paris 6)\\
\url{sothea.has@lpsm.paris}
\end{flushleft}
\begin{abstract}
 In this paper, we present a study of a kernel-based consensual aggregation on randomly projected high-dimensional features of predictions for regression. The aggregation scheme is composed of two steps: the high-dimensional features of predictions, given by a large number of regression estimators, are randomly projected into a smaller subspace using Johnson-Lindenstrauss Lemma in the first step, and a kernel-based consensual aggregation is implemented on the projected features in the second step. We theoretically show that the performance of the aggregation scheme is close to the performance of the aggregation implemented on the original high-dimensional features, with high probability. Moreover, we numerically illustrate that the aggregation scheme upholds its performance on very large and highly correlated features of predictions given by different types of machines. The aggregation scheme allows us to flexibly merge a large number of redundant machines, plainly constructed without model selection or cross-validation. The efficiency of the proposed method is illustrated through several experiments evaluated on different types of synthetic and real datasets.
\end{abstract}

\bigskip


\bigskip


\section{Introduction}

In supervised machine learning problems, one aims at predicting values of any quantities of interest using the corresponding input information. When the quantity of interest or {\it response} takes continuous values (which is the focus of this paper), the task is called {\it regression}. On the other hand, it is called {\it classification} if the response takes values in any finite sets (few unique values).

Nowadays, several machine learning models are invented, can be easily implemented and used in any supervised prediction problems. Those methods aim at approximating the relationship between inputs and the corresponding outputs by minimizing some empirical criterion, which is a function of the training data. Hence, the performances of those predictive models strongly depend on the data fed to them. In practice, we may try to implement different types of models according to the context of the problems, and the one with strong generalization capability would be selected. However, selecting the best method may require a lot of efforts and consideration. Therefore, another approach is to automatically combine those candidate predictors in a flexible way, in a sense that the performance of the combination biases towards the best basic estimators.

Up to now, many combining estimation methods have been introduced, for instance, ensemble learning methods which combines an homogeneous type (trees) of predictors such as Random Forest (\cite{baggingPredictors}) and Boosting (\cite{gradientBoosting}). Moreover, some other methods allowing to combine a bunch of different types of individual estimators using some convex combination are also introduced, for example, in \cite{Cat04}, \cite{JN2000}, \cite{Nem00}, \cite{yang2000,yang2001}, \cite{yang2004}, \cite{bookDistributionFree}, \cite{W03}, \cite{Aud}, \cite{BTW06,BTW07a,BTW07b}, and \cite{DalTsy}. There are also a group of combining strategies that aggregate different instance estimators based on features of predictions given by the basic estimators such as stack generalization of \cite{WOLPERT1992241} and stacked regression by \cite{Breiman1996StackedReg}. Last but not least, some combining estimation methods aggregating different types of individual estimators based on consensus level of predictions given by the instances, which is the central idea of this chapter, are also introduced by \cite{mojirsheibani1999, mojirsheibani2000} and \cite{majidAndKong2016} for classification problems, by \cite{cobra} and \cite{has2021} for regression problems, and for both frameworks by \cite{mixcobra}, where in this last method the combination also takes into account the input part. The consistency result of each consensual aggregation method is provided under different assumptions, and is also confirmed through several numerical simulations.

This study focuses on a high-dimensional setting of combining estimation strategy for regressions by \cite{has2021}. The method is an extension to a regular kernel-based framework of a combining strategy by \cite{cobra}, which is a regression configuration of combining classifiers by \cite{mojirsheibani1999}. More precisely, let ${\bf r}(x)=(r_1(x),...,r_M(x))$ denote the prediction vector of $x\in\mathbb{R}^d$, given by the $M$ basic regression estimators $r_1,...,r_M$, and suppose that $n$ iid couples of supervised training data $(X_1,Y_1),...,(X_n,Y_n)$ are observed. Moreover, let $\|.\|$ denote the Euclidean norm on $\mathbb{R}^M$, thus the prediction at any point $x\in\mathbb{R}^d$ of the combining strategy by \cite{has2021} is defined by
\begin{equation}
\label{eq:KCOBRA}
g_n(\textbf{r}(x))=\frac{\sum_{i=1}^nY_iK_{h}(\|\textbf{r}(x)-\textbf{r}(X_i)\|)}{\sum_{j=1}^nK_{h}(\|\textbf{r}(x)-\textbf{r}(X_j)\|)}
\end{equation}
for some regular kernel function $K$ with $K_h(x)=K(x/h)$ for some smoothing parameter $h>0$, and the convention of $0/0=0$. Note that COBRA method of \cite{cobra} corresponds to naive kernel $K(x)=\prod_{j=1}^M\mathbbm{1}_{\{|x_j|<\varepsilon\}}$ for some window parameter $\varepsilon>0$ to be tuned. It is theoretically shown that the combining strategy asymptotically outperforms the best individual estimator in $L_2$ sense. Moreover, the implementation of the classical method is available in \texttt{COBRA} library of \texttt{R} software (see~\cite{cobraR}), and a slightly different setting of its kernel-based configuration is available in Python library called \texttt{pycobra} (see~\cite{pycobra}). 

Until now, the study of high-dimensional case of the described consensual aggregation method has not been considered yet. Therefore, this study aims at filling this gap by considering exponential kernel-based consensual aggregation for regression on high-dimensional features of predictions. In other words, we are interested in combining a large number of basic machines, which might be obtained by varying the hyperparameters of any types of predictive models, or from mixtures of different types of models. Moreover, these basic machines can be constructed without any model selection or cross-validation techniques. One can simply see this aggregation scheme as a method to merge the candidate models into one final prediction that is asymptotically optimal with respect to all the basic machines. 

However, working in high-dimensional spaces often brings along some difficulties such as highly computational cost and curse of dimensionality, which refers to the situation where Euclidean distance loses its meaning. In this study, these problems are handled using dimensional reduction technique based on Johnson and Lindenstrauss Lemma (J-L). Johnson and Lindenstrauss showed that for any $\delta>0$ given, one can embed a given finite set of high-dimensional vectors of Euclidean spaces into a lower-dimensional subspace, preserving the pairwise Euclidean distances between data points up to an error $\delta$, with high probability (see, for example, \cite{JL1984} and \cite{JL1986}). This result has become a very powerful technique of dimensional reduction aiming at preserving pairwise Euclidean distances between data points (\cite{JLGraphs1988,someGeoApplic1990} and \cite{AnElementaryProof2003}). J-L method is suitable for our setting not only because of the pairwise-distance preserving property, but also because of its computational efficiency. The implementation of this method is as simple as simulating $M$ independent random vectors (rows of projection matrix), and performing a matrix multiplication. Dimensional reduction based on J-L technique has also been applied in several machine learning studies, for instance, in image processing and text analysis by \cite{RandomProjectImageText}, in Lipschitz embeddings of graphs into normed spaces by \cite{JLGraphs1988}, in approximating nearest-neighbor in high-dimensional spaces by \cite{TwoAlgoNN} and \cite{ApproxNN}, in linear regression framework by \cite{lmWithJL}, and also in unsupervised clustering in Hilbert spaces by \cite{onThePerformanceofJL}.


In this work, we propose an aggregation scheme on random projected features of high-dimensional predictions. The scheme is composed of two steps. First, we randomly embed the original features of predictions of dimension $M$ (large) into a lower subspace of dimension $m$ ($m<M$) using dimensional reduction based on J-L Lemma. Then, the consensual aggregation \eqref{eq:KCOBRA} is implemented on the projected features of predictions in the second step. We aim in this study to provide a probability bound of the difference between the classical consensual aggregation and the aggregation implemented on projected features of predictions. We also numerically illustrate the performance of the full aggregation scheme on several simulated and real-world datasets.

This chapter is organized in the following manner. Section~\ref{sec:ch4method} details the construction of the proposed aggregation scheme. Section~\ref{sec:ch4theory} provides the theoretical performance of the method. Section~\ref{sec:ch4numeric} illustrates performance of the method through several numerical experiments evaluated on different types of datasets. Lastly, the proofs of the theoretical results stated in this paper are collected in Section~\ref{sec:ch4proof}.
\section{The aggregation method}
\label{sec:ch4method}
\subsection{Notation}
Assume that $(X,Y)$ is an $\mathbb{R}^d\times\mathbb{R}$-valued generic random variable, and that we have at hand a training dataset containing iid copies of $(X,Y)$:
$${\cal D}_n=\{(X_1,Y_1),(X_2,Y_2),...,(X_n,Y_n)\}.$$
We assume moreover that $M$ basic regression estimators or machines $r_1, r_2,$ $..., r_M$, are constructed independently of $D_n$ (otherwise, a simple splitting technique can be used as described, for example, in \cite{cobra} and \cite{has2021}). These basic machines can be any regression estimators of the same type (with different parameters), or constructed based on completely different theories. We only require that they can predict the training data and any new data points since the aggregation is done based only on those predictions.

To alleviate notation, when the context is clear, all Euclidean norms will be denoted by $\|.\|$ without mentioning the dimension of the space. Moreover, this paper deals with exponential kernel, $K(t)=\exp(-t^{\alpha}/\sigma)$, for some $\sigma>0$ and $\alpha\geq0$, which has numerically been shown to be the most outstanding one so far in the previous studies. Moreover, let $\mu$ denote the distribution of $X$ with respect to Lebesgue measure, and the regression function is denoted by $\eta(x)=\mathbb{E}[Y|X=x]$. 

\subsection{Random projection: Johnson-Lindenstrauss Lemma}
In the sequel, the prediction matrix of the training data is denoted by

\begin{equation}
\label{eq:projFeatures}
\textbf{r}(\mathcal{X})=\begin{pmatrix}
  r_1(X_1) &  r_2(X_1) & \hdots &  r_M(X_1) \\
  r_1(X_2) &  r_2(X_2) & \hdots &  r_M(X_2) \\
  \vdots & \vdots & \vdots & \vdots \\
  r_1(X_n) &  r_2(X_n) & \hdots &  r_M(X_n) \\
\end{pmatrix}_{n \times M}.
\end{equation} 
For any positive integer $m<M$, let $G=(G_{ij})_{1\leq i\leq M,1\leq j\leq m}$ be a {\it random projection} matrix where the entries $G_{ij}$ are iid centered Gaussian random variables with variance $1/m$, for all $i=1,2,...,M$ and $j=1,2,...,m$. Embedding the predicted features \eqref{eq:projFeatures} into a subspace of dimension $m$ via J-L random projection is simply done by multiplying the matrix of original features $\textbf{r}(\mathcal{X})$ by a random projection matrix $G$ i.e.,
\begin{align*}
\tilde{\textbf{r}}(\mathcal{X})&=\textbf{r}(\mathcal{X})\times G\\
&=\begin{pmatrix}
  r_1(X_1)  & \hdots &  r_M(X_1) \\
  \vdots & \ddots & \vdots \\
  r_1(X_n) & \hdots &  r_M(X_n) \\
\end{pmatrix}\times
\begin{pmatrix}
  G_{11}  & \hdots &  G_{1m} \\
  \vdots & \ddots & \vdots \\
  G_{M1}  & \hdots &  G_{Mm} \\
\end{pmatrix}\\
&=\begin{pmatrix}
  \tilde{r}_1(X_{1})  & \tilde{r}_2(X_{1})   & \hdots &  \tilde{r}_m(X_{1})  \\
  \tilde{r}_1(X_{2})  & \tilde{r}_2(X_{2})   & \hdots &  \tilde{r}_m(X_{2})  \\
  \vdots & \vdots & \vdots & \vdots \\
  \tilde{r}_1(X_{n})  & \tilde{r}_2(X_{n})   & \hdots &  \tilde{r}_m(X_{n})  \\
\end{pmatrix}_{n \times m}.
\end{align*}
The $i$th row-vector of $\tilde{\bf r}(\mathcal{X})$ is the vector of embedded features evaluated at $X_i$, denoted by $\tilde{\bf r}(X_i)=(\tilde{r}_1(X_i),\tilde{r}_2(X_i), ..., \tilde{r}_m(X_i))$ for $i=1,2, ...,n$. It is easy to check that given the original features ${\bf r}(X_i)$ and ${\bf r}(X_j)$, the Euclidean distance between its projection $\|\tilde{\bf r}(X_i)-\tilde{\bf r}(X_j)\|$, is equal to the Euclidean distance between the original pair $\|{\bf r}(X_i)-{\bf r}(X_j)\|$ in expectation with respect to $G$. More precisely, since $G_{ij}$ are centered and iid, one has

\begin{align*}
&\quad \mathbb{E}_{\mathcal{G}}[\|\tilde{\bf r}(X_i)-\tilde{\bf r}(X_j)\|^2|{\bf r}(X_i),{\bf r}(X_j)]\\
&=\sum_{p=1}^m\mathbb{E}_{\mathcal{G}}[(\tilde{r}_p(X_i)-\tilde{r}_p(X_j))^2|{\bf r}(X_i),{\bf r}(X_j)]\\
&=\sum_{p=1}^m\mathbb{E}_{\mathcal{G}}\Big[\Big(\sum_{k=1}^M(r_k(X_i)-r_k(X_j))G_{kp}\Big)^2|{\bf r}(X_i),{\bf r}(X_j)\Big]\\
&=\sum_{p=1}^m\sum_{k=1}^M(r_k(X_i)-r_k(X_j))^2\mathbb{E}_{\mathcal{G}}[G_{kp}^2|{\bf r}(X_i),{\bf r}(X_j)] &(\mathbb{E}_{\cal G}[G_{kp}]=0)\\
&=\sum_{p=1}^m\sum_{k=1}^M(r_k(X_i)-r_k(X_j))^2/m &(\mathbb{E}_{\cal G}[G_{kp}^2]=1/m)\\
&=\sum_{k=1}^M(r_k(X_i)-r_k(X_j))^2=\|{\bf r}(X_i)-{\bf r}(X_j)\|^2,
\end{align*}
where $\mathbb{E}_{\cal G}$ denotes the expectation with respect to $G$. Moreover, as the $p$th coordinate of vector $\tilde{\bf r}(X_i)-\tilde{\bf r}(X_j)$ is given by

\begin{align*}
(\tilde{\bf r}(X_i)-\tilde{\bf r}(X_j))_p&=\tilde{r}_p(X_i)-\tilde{r}_p(X_j)=\sum_{k=1}^M(r_k(X_i)-r_k(X_j))G_{kp},
\end{align*}
and one has
\begin{equation*}
(\tilde{\bf r}(X_i)-\tilde{\bf r}(X_j))_p\sim{\cal N}(0,\|{\bf r}(X_i)-{\bf r}(X_j)\|^2/m),\text{ for all } p=1,2,...,m.
\end{equation*}
Therefore,
\begin{equation*}
m\frac{\|\tilde{\bf r}(X_i)-\tilde{\bf r}(X_j)\|^2}{\|{\bf r}(X_i)-{\bf r}(X_j)\|^2}\sim\mathcal{\chi}^2(m).
\end{equation*}
Then, the gap between the original and projected features can by controlled using concentration inequalities, for example, by applying Chernoff bound for $\mathcal{\chi}^2(m)$ distribution (see \cite{Chernoff2011}), for any rows ${\bf r}(X_i)$ and ${\bf r}(X_j)$ of $\textbf{r}(\mathcal{X})$, and for any $\delta>0$, one has
\begin{equation}
\label{eq:JL1}
\mathbb{P}_{\mathcal{G}}\Big(\frac{\|\tilde{\bf r}(X_i)-\tilde{\bf r}(X_j)\|^2}{\|{\bf r}(X_i)-{\bf r}(X_j)\|^2}-1>\delta\Big)\leq e^{m[-\delta+\ln(1+\delta)]/2}
\end{equation}
and
\begin{equation}
\label{eq:JL2}
\mathbb{P}_{\mathcal{G}}\Big(\frac{\|\tilde{\bf r}(X_i)-\tilde{\bf r}(X_j)\|^2}{\|{\bf r}(X_i)-{\bf r}(X_j)\|^2}-1<-\delta\Big)\leq e^{m[\delta+\ln(1-\delta)]/2},
\end{equation}
where $\mathbb{P}_{\cal G}$ denotes the probability under the law of $G$. The union bound of inequalities~\eqref{eq:JL1} and \eqref{eq:JL2}, together with the following inequalities
\begin{equation}
\label{eq:inequality}
\begin{cases}
\ln(1+\delta)&\leq \delta-\frac{\delta^2}{2}+\frac{\delta^3}{3}\\
\ln(1-\delta)&\leq -\delta-\frac{\delta^2}{2}-\frac{\delta^3}{3}
\end{cases},
\end{equation}
for any $\delta\in (0,1)$, yields the following proposition.
\begin{prop}(Johnson-Lindenstrauss)
\label{prop:JL1}
Let $S_n = \{z_j\in\mathbb{R}^M:j=1,2,...,n\}$ denote a subset containing $n$ points of $\mathbb{R}^M$ and $z_0\in\mathbb{R}^M$ fixed. Moreover, let $\tilde{z_0}$ and $\tilde{z_j}$ denote the projected point of $z_0$ and $z_j$ respectively into $\mathbb{R}^m$ using random projection described above. Thus, for any $\delta\in(0,1)$, with probability at least $1-2n\exp(-m(\delta^2/2-\delta^3/3)/2)$, one has: 
$$\Big|\frac{\|\tilde{z_0}-\tilde{z_j}\|^2}{\|z_0-z_j\|^2}-1\Big|\leq \delta,\text{ for all } z_j\in S_n.$$
\end{prop}
\subsection{Aggregation on random projected features}
We are now in a position to formally describe our aggregation strategy on random projected features of high-dimensional predictions. We first embed the original $M$-dimensional features of predictions ${\bf r}(\cal X)$ using J-L random projection, simply by multiplying ${\bf r}(\cal X)$ by a random projection matrix $G$ to obtain the projected features $\tilde{\bf r}(\cal X)$. Then, the aggregation method \eqref{eq:KCOBRA} is implemented on the projected features $\tilde{\bf r}(\cal X)$ in the last step. More precisely, the prediction of any point $x\in\mathbb{R}^d$ is defined by
\begin{equation}
\label{eq:RPKCOBRA}
g_n(\tilde{\bf r}(x))=\frac{\sum_{i=1}^nY_iK_h(\|(\tilde{\textbf{r}}(x)-\tilde{\textbf{r}}(X_i)\|)}{\sum_{j=1}^nK_h(\|\tilde{\textbf{r}}(x)-\tilde{\textbf{r}}(X_j)\|)}.
\end{equation}
Note that for any $x\in\mathbb{R}^d$ one has $\tilde{\textbf{r}}(x)\in\mathbb{R}^m$ and the Euclidean norm used in \eqref{eq:RPKCOBRA} is defined on $\mathbb{R}^m$ while the one used in \eqref{eq:KCOBRA} is defined on $\mathbb{R}^M$.
\section{Theoretical performance}
\label{sec:ch4theory}
In the sequel, we assume that dimension $M$ of the predicted features is large. Moreover, the consensual aggregation method implemented on the original $M$-dimensional features of predictions (respectively $m$-dimensional projection features) is called \textit{full} (respectively \textit{projected}) aggregation method. 

We are now in a position to state the main theoretical result regarding the difference between the full and projected aggregation methods. More precisely, for any $\varepsilon>0$, we are interested in controlling the following probability:
\begin{equation}
\label{eq:errorProject}
\mathbb{P}\Big(g_n({\bf r}(X))-g_n(\tilde{\bf r}(X))|>\varepsilon\Big)
\end{equation}
where $g_n({\bf r}(.))$ and $g_n(\tilde{\bf r}(.))$ are the two aggregation methods defined respectively in \eqref{eq:KCOBRA} and \eqref{eq:RPKCOBRA}. The key difference between the two methods is the features of predictions used for the aggregation, therefore the proof relies on the theoretical result of J-L Lemma. The control of this probability is given in the following theorem.
\begin{theorem}
\label{prop:question1}
Assume that all the machines $r_1,r_2,...,r_M$ and the response variable $Y$ are bounded almost surely by $R_0$, thus for any $h,\varepsilon>0, n\geq1,$ and for any $\delta\in(0,1)$, with the choice of $m$ satisfying: $$m\geq C_1\frac{\log[2/(1-\sqrt[n]{1-\delta})]}{h^{2\alpha}\varepsilon^2},\ \text{with }C_1=3(2+\alpha)^2(2R_0)^{2(1+\alpha)}/\sigma^2,$$ one has:
\begin{align*}
\mathbb{P}\Big(|g_n(\textbf{r}(X))-g_n(\tilde{\textbf{r}}(X))|>\varepsilon\Big)\leq \delta.
\end{align*}
\end{theorem}
The probability of Theorem~\ref{prop:question1} is computed under the laws of $X$, the training data $\mathcal{D}_n=\{(X_i,Y_i)_{i=1}^n\}$ and the random projection matrix $G$. It can be viewed as the loss of aggregation method when projecting the features of predictions into smaller subspace of dimension $m$. Note that in this result, the constant $C_1$ depends on $R_0$, which is in practice can be scaled to be, for example, less then $1$. Therefore, the constant $C_1\approx 12$ for Gaussian kernel, and the lower bound of $m$ is roughly of order: $$O\Big(\frac{\log(2n/\delta)}{\varepsilon^2h^{2\alpha}}\Big),$$
for large $n$ and small $\delta$.

\section{Numerical simulation}
\label{sec:ch4numeric}
This section is devoted to numerical experiments carried out on several simulated and real datasets to illustrate the performance of the proposed method. The basic regression machines considered in this section are of five different types:
\begin{itemize}
\item \textcolor{cyan}{\bf kNN}: $k$-nearest neighbors for regression (R package \texttt{FNN}, see \cite{FNNR}).
\item \textcolor{cyan}{\bf Elas}: lasso and elastic-net regularized generalized linear models (R package \texttt{glmbet}, see \cite{glmnetR})
\item \textcolor{cyan}{\bf Bag}: bagging tree for regression (R package \texttt{ipred}, see \cite{baggingR}).
\item \textcolor{cyan}{\bf RF}: regression random forest (R package \texttt{randomForest}, see \cite{randomForestR1}).
\item \textcolor{cyan}{\bf Boost}: gradient boosting (R package \texttt{gbm}, see \cite{gbmR}).
\end{itemize}
To produce high-dimensional features of predictions, we construct the basic machines of each type using various options of the corresponding parameters as described below:
\begin{itemize}
\item $200$ values of $k\in\{2,3,...,201\}$ for \textcolor{cyan}{\bf kNN}.
\item The coefficients of elastic-net model are defined by
$$\hat{\beta}=\argmin_{\beta}\{\|Y-\beta X\|_2^2+\lambda[\alpha\|\beta\|_1+(1-\alpha)\|\beta\|_2^2]\},$$
where $\alpha$ is the trade-off parameter between $L_1$ and $L_2$ penalty, and $\lambda$ is the penalty parameter. In this case, $5\times100=500$ values of the couple $(\alpha,\lambda)\in\{0,0.25,0.5,0.75,1\}\times\{0.00005,...,1\}$ are considered. Note that $\alpha=0$ (respectively $\alpha=1$) corresponds to \textcolor{cyan}{\bf Ridge} (respectively \textcolor{cyan}{\bf Lasso}) regression. 
\item $100$ values of $ntree\in\{18,21,...,315\}$ for the three remaining tree-based methods: \textcolor{cyan}{\bf Bag}, \textcolor{cyan}{\bf RF} and \textcolor{cyan}{\bf Boost}.
\end{itemize}
\begin{remark}
With the choices of parameters of each model, one may expect the features of predictions to be very highly correlated or redundant. For example, many values of parameter $k$ of \textcolor{cyan}{\bf kNN}, and $ntree$ of \textcolor{cyan}{\bf Bag} and \textcolor{cyan}{\bf RF} are not very interesting in a normal setting, however, in our context, it is quite interesting to see the performance of the aggregation method in such a large highly correlated features. This is interesting in a sense that, without model selection or cross-validation technique, the aggregation method can merge the features of predictions in a robust way.
\end{remark}
Therefore, the features of predictions are of dimension $1000$. The performance of any regression estimator $f$ is measured using the following {\it root mean square error} (RMSE) evaluated on an independent testing dataset: 
$$\text{RMSE}(f)=\sqrt{\frac{1}{n_\text{test}}\sum_{i=1}^{n_\text{test}}(f(x_i)-y_i)^2}$$
where $n_{\text{test}}$ denotes the number of testing sample. 
\subsection{Simulated datasets}
In this part, we consider 5 simulated models of size $n$ where the $d$-dimensional input data is uniformly distributed on $[-1,1]^d$, denoted by $X\sim\mathcal{U}([-1,1]^d)$. The five simulated models are defined as follows:
\begin{model}{:}
\label{mod:1}
$n=600, d=10,$\\ $\displaystyle Y=X_1^2-X_3^2+3X_4\exp(-X_5)-X_7^3\exp(-X_8X_9+X_5X_{10})+\mathcal{N}(0,1).$
\end{model}
\begin{model}{:}
\label{mod:2}
$n=800, d=30,$\\ $\displaystyle Y=\sum_{j=1}^5[3X_{2j}^3\exp(X_{30-j}-X_{2j+1}) - 2X_{2j-1}^3\exp(X_{2j}-X_{30-3j})]+\mathcal{N}(0,1).$
\end{model}
\begin{model}{:}
\label{mod:3}
$n=800, d=50,$\\ $\displaystyle Y=\frac{1-X_{1}^2+2X_{3}X_{4}}{1.1+X_{5}}-2\sqrt{1+\sum_{j=1}^5\frac{1+X_{5+j}}{2-X_{45+j}}}\exp(-X_{10}+X_{20}-X_{30})+\mathcal{N}(0,1).$
\end{model}
\begin{model}{:}
\label{mod:4}
$n=800, d=100,$\\ $\displaystyle Y=(X_1^2-X_2^2)(1-\exp(-X_5X_7))+3X_3\exp(-\sum_{j=1}^{10}X_{10j})+\mathcal{N}(0,1).$
\end{model}
\begin{model}{:}
\label{mod:5}
$n=800, d=100,$\\ $\displaystyle Y=\frac{1+\sin(X_1+X_2)}{1-\sin(X_1X_2)}-\sum_{j=1}^{10}\frac{2^{j}+1}{2^{j}-1}X_{5j}X_{10j}X_{j}+\mathcal{N}(0,1).$
\end{model}
In each simulation, we randomly split the simulated data into $80\%$ and $20\%$ training and testing set respectively. Then, the training data is split further into two parts of sizes $n_1$ and $n_2$ such that $n_1=\lceil n_{\text{train}}/2\rceil=n_{\text{train}}-n_2$. The first part of the training data of size $n_1$ is used to construct the $1000$ machines yielding predictions of the remaining parts. On top of that, to study the impact of the projected dimension $m$, the matrix of original features of predictions ${\bf r}({\cal X})$ is embedded into two groups of subspaces. The first group corresponds to the case of $m\in\{100,200,...,900\}$, and the second group consists of much smaller values of $m\in\{2,3,...,9\}$, associated with different random projection matrices $G$. Then, the kernel-based consensual aggregation method of equation~\eqref{eq:RPKCOBRA} is implemented. Moreover, the aggregation on the original features defined in equation~\eqref{eq:KCOBRA} is also computed and used to compare with all the projected cases. 

\begin{sidewaystable}[!htbp]
\scriptsize
\centering  
\hspace{0.5em}
\setlength{\tabcolsep}{3.5pt}
\def\arraystretch{1.2}
\begin{tabular}{ c | c c c c c | c c c c c c c c c }  
\hline\hline                       
\multirow{2}{*}{\textcolor{cyan}{\bf Model}} & \multicolumn{5}{c|}{\bf Basic machines} & \multicolumn{9}{c}{{\bf Aggregation method} \textcolor{cyan}{\bf Comb$m$}}\\ [0.5ex] \cline{2-15}
 & \textcolor{cyan}{\bf $k$NN} &\textcolor{cyan}{\bf Elas} &\textcolor{cyan}{\bf Bag} & \textcolor{cyan}{\bf RF} & \textcolor{cyan}{\bf Boost} & \textcolor{cyan}{\bf 100/2} & \textcolor{cyan}{\bf 200/3} & \textcolor{cyan}{\bf 300/4} & \textcolor{cyan}{\bf 400/5}  & \textcolor{cyan}{\bf 500/6} & \textcolor{cyan}{\bf 600/7} & \textcolor{cyan}{\bf 700/8} & \textcolor{cyan}{\bf 800/9} & \textcolor{cyan}{\bf 900/Comb\_Full}\\ [0.5ex]   
\hline  
\multirow{4}{*}{\ref{mod:1}} & \multirow{2}{*}{\raisebox{-2.5ex}{1.620}} & \multirow{2}{*}{\raisebox{-2.5ex}{1.579}} & \multirow{2}{*}{\raisebox{-2.5ex}{1.241}} & \multirow{2}{*}{\raisebox{-2.5ex}{1.304}} & $\multirow{2}{*}{\raisebox{-2.5ex}{\bf 1.116}}$ & {\bf 1.081} & 1.083 & 1.083 & 1.082 & 1.083 & {\bf 1.081} & 1.083 & 1.082 & 1.084 \\
& & & & & & $(0.030)$ & $(0.033)$ & $(0.032)$ & $(0.032)$ & $(0.033)$ & $(0.031)$ & (0.032) & (0.033) &  (0.032)\\ \cline{7-15}
& (0.102) & (0.091)& (0.064) & (0.087) & (0.071) &  1.152 & 1.106  & 1.092  & 1.095 & 1.097 & 1.092 & 1.092 & 1.086 & {\bf 1.083} \\ 
 & & & & & & $(0.064)$ & $(0.038)$ & $(0.034)$ & $(0.037)$ & $(0.038)$ & $(0.038)$ & (0.036) & (0.038) & (0.032) \\
 \hline 
 \multirow{4}{*}{\ref{mod:2}} & \multirow{2}{*}{\raisebox{-2.5ex}{4.498}} & \multirow{2}{*}{\raisebox{-2.5ex}{3.971}} & \multirow{2}{*}{\raisebox{-2.5ex}{4.203}} & \multirow{2}{*}{\raisebox{-2.5ex}{4.081}} & $\multirow{2}{*}{\raisebox{-2.5ex}{\bf 3.621}}$ & {\bf 3.413} & 3.425 & 3.423 & 3.429 & 3.419 & 3.417 & 3.428 & 3.423 & 3.416\\
 & & & & & & (0.138) & (0.145) & (0.145) & (0.140) & (0.142) & (0.151) & (0.132) & (0.137) & (0.152)\\ \cline{7-15}
&(0.314) & (0.275) & (0.298) & (0.293) & (0.269) & 3.441 & 3.474 & 3.411 & 3.445 & 3.412 & 3.437 & 3.429 & {\bf 3.400} & 3.427\\
& & & & & & $(0.139)$ & $(0.142)$ & $(0.134)$ & $(0.168)$ & $(0.171)$ & $(0.167)$ & (0.149) & (0.150) & (0.138) \\
  \hline  
 \multirow{2}{*}{\ref{mod:3}} & \multirow{2}{*}{\raisebox{-2.5ex}{5.525}} & \multirow{2}{*}{\raisebox{-2.5ex}{4.037}} & \multirow{2}{*}{\raisebox{-2.5ex}{3.144}} & \multirow{2}{*}{\raisebox{-2.5ex}{3.454}} & $\multirow{2}{*}{\raisebox{-2.5ex}{\bf 2.518}}$ & 2.038 & 2.035 & 2.264 & {\bf 2.028} & 2.037 & 2.040 & 2.145 & 2.041 & 2.031\\
& & & & & & (0.126) & (0.135) & (0.855) & (0.130) & (0.141) & (0.132) & (0.582) & (0.140) & (0.127)\\ \cline{7-15}
&(0.768) & (0.584) & (0.382) & (0.526) & (0.333) & 2.116 & 2.124 & 2.173 & 2.060 & 2.072 & 2.070 & 2.082 & 2.060 & {\bf 2.044}\\
& & & & & & (0.150) & (0.181) & (0.619) & (0.160) & (0.163) & (0.146) & (0.166) & (0.152) & (0.131)\\
 \hline  
 \multirow{2}{*}{\ref{mod:4}} & \multirow{2}{*}{\raisebox{-2.5ex}{18.752}} & \multirow{2}{*}{\raisebox{-2.5ex}{18.350}} & \multirow{2}{*}{\raisebox{-2.5ex}{17.844}} & \multirow{2}{*}{\raisebox{-2.5ex}{$ 18.706$}} & $\multirow{2}{*}{\raisebox{-2.5ex}{\bf 17.708}}$ & 15.672 & 15.677 & 15.610 & 15.785 & {\bf 15.573} & 15.822 & 15.814 & 15.741 & 15.604\\
& & & & & & (3.566) & (3.488) & (3.528) & (3.532) & (3.536) & (3.449) & (3.753) & (3.539) & (3.564)\\ \cline{7-15}
 & (4.847) & (4.626) & (4.497) & (4.409) & (4.632) & 16.962 & 16.823 & 16.914 & 16.210 & 16.362 & 16.142 & 16.150 & 16.092 & {\bf 15.745}\\
& & & & & & (4.993) & (5.049) & (4.912) & (4.889) & (4.723) & (4.874) & (4.963) & (4.976) & (3.609)\\
   \hline  
 \multirow{2}{*}{\ref{mod:5}} & \multirow{2}{*}{\raisebox{-2.5ex}{1.417}} & \multirow{2}{*}{\raisebox{-2.5ex}{1.169}} & \multirow{2}{*}{\raisebox{-2.5ex}{\bf 1.021}} & \multirow{2}{*}{\raisebox{-2.5ex}{1.076}} & $\multirow{2}{*}{\raisebox{-2.5ex}{1.031}}$ & 0.955 & 0.955 & {\bf 0.953} & 0.956 & 0.955 & 0.955 & 0.956 & 0.956 & {\bf 0.953}\\
 & & & & & & (0.039) & (0.043) & (0.040) & (0.042) & (0.040) & (0.044) & (0.040) & (0.041) & (0.042)\\ \cline{7-15}
&(0.114) & (0.086) & (0.046) & (0.068) & (0.045) & 0.950 & 0.951 & 0.948 & {\bf 0.942} & 0.956 & 0.953 & 0.950 & 0.953 & 0.954\\
 & & & & & & (0.054) & (0.057) & (0.067) & (0.048) & (0.057) & (0.050) & (0.050) & (0.050) & (0.041)\\
 \hline                             
  \hline 
\end{tabular}  
\caption{Average RMSEs on all simulated datasets.}
\label{tab:simulated}
\end{sidewaystable}

The average RMSE and the associated standard error (into bracket) over $30$ independent runs of each model are reported in Table~\ref{tab:simulated} below. For the sake of readability, only the best performance of each type of the five basic machines is reported, followed by the performance of all the aggregation methods. 
In this table, the first block consists of five columns (2nd to 6th), corresponding to the performances of the best cases of the five basic machines (\textcolor{cyan}{\bf $k$NN}, \textcolor{cyan}{\bf Elas}, \textcolor{cyan}{\bf Bag}, \textcolor{cyan}{\bf RF} and \textcolor{cyan}{\bf Boost}), and the second block contains 9 columns (two rows in each column) corresponding to the results of the aggregation method with different values of $m$. The column's names of this block are of the form \textcolor{cyan}{\bf $m_1/m_2$}, where \textcolor{cyan}{\bf $m_1$} and \textcolor{cyan}{\bf $m_2$} are the dimensions of the projected subspaces reported in the first and second row respectively (except for the last column \textcolor{cyan}{\bf 900/Comb\_Full}). More precisely, the first row of this block contains the results of the projected aggregation methods with $m\in\{100, 200,....900\}$, and the second row contains the performances of the methods with $m=2,3,...,9$, plus the full aggregation method, which is the aggregation on the original predicted features of dimension $M=1000$ (the second row of the last column). In each case, the best performance of each block is written in {\bf boldfaced}.  

\begin{figure}[!htbp]
\centering
\includegraphics[width=0.49\linewidth]{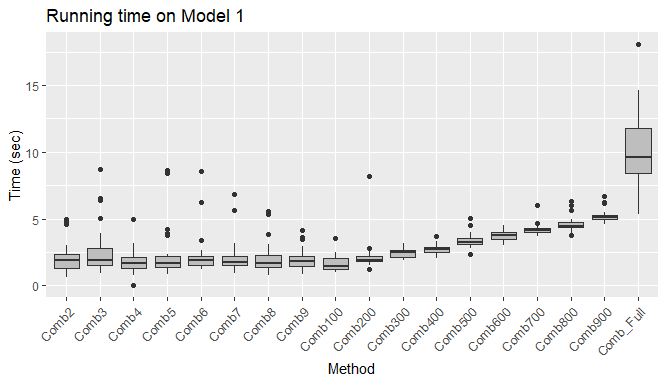}
\includegraphics[width=0.49\linewidth]{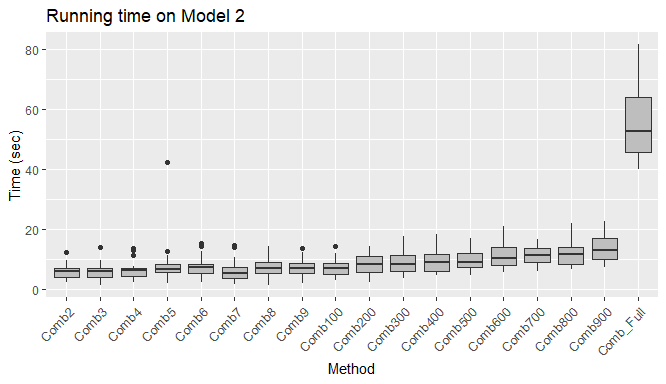}
\includegraphics[width=0.49\linewidth]{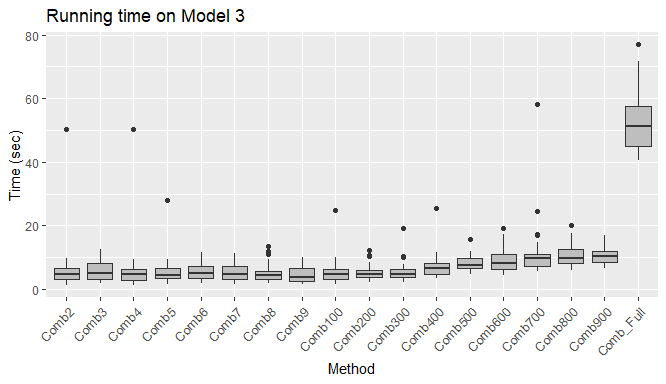}
\includegraphics[width=0.49\linewidth]{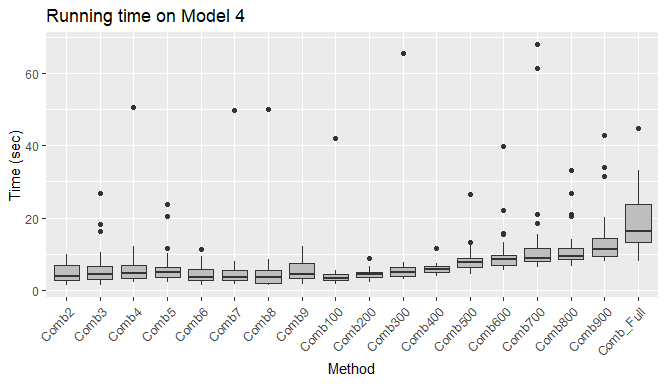}
\includegraphics[width=0.49\linewidth]{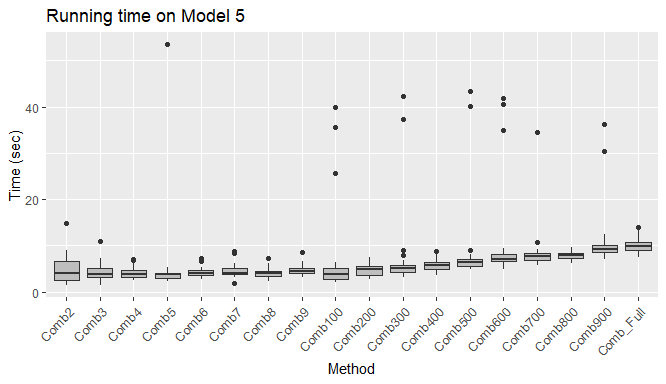}
\caption{Running times of all the combining methods on simulated datasets. With approximately the same accuracy, the proposed methods are at least $3$ times faster than the full aggregation.}
\label{fig:boxTimeSimulated}
\end{figure}

We observe in Table~\ref{tab:simulated} that \textcolor{cyan}{\bf Boost} shows the best performance comparing to other basic machines in the first block. In the second blocks, we see that the performances of all aggregation methods are quite similar which confirms the theoretical result stated in Theorem~\ref{prop:question1}. Moreover, the performances of the aggregations bias towards, sometimes even outperform, the best method of the first block. We can also see that the full aggregation method (second row of the last column) performs really well despite being implemented on a very large redundant set of machines. And more interestingly, the performances of all the proposed methods are preserved in much lower dimensional spaces (second rows of the second block). In addition to that, Figure~{\ref{fig:boxTimeSimulated}} below provides the computational efficiency of the method implemented using a computational machine with the following characteristics:
\begin{itemize}
\item Processor: 2x AMD Opteron 6174, 12C, 2.2GHz, 12x512K L2/12M L3 Cache, 80W ACP, DDR3-1333MHz.
\item Memory: 64GB Memory for 2 CPUs, DDR3, 1333MHz.   
\end{itemize}

\begin{figure}[!htbp]
\centering
\includegraphics[width=0.49\linewidth]{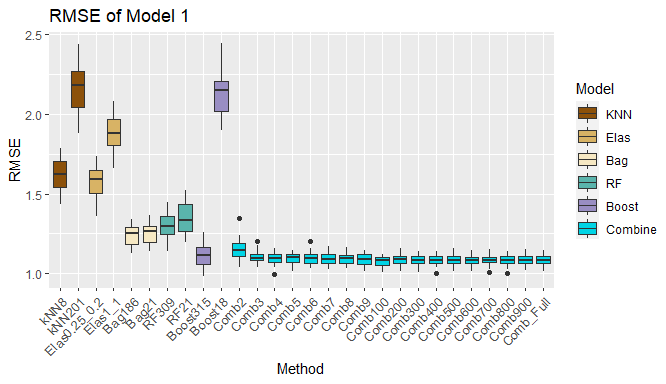}
\includegraphics[width=0.49\linewidth]{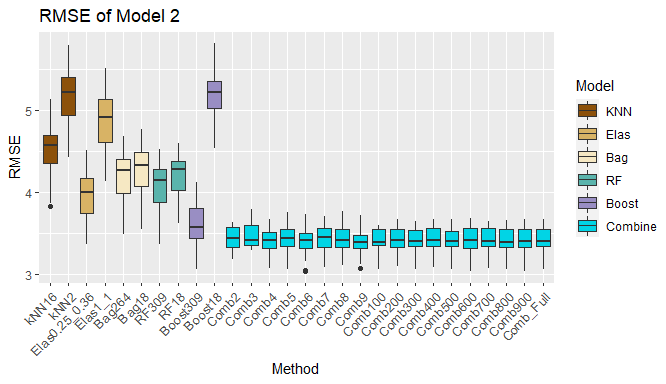}
\includegraphics[width=0.49\linewidth]{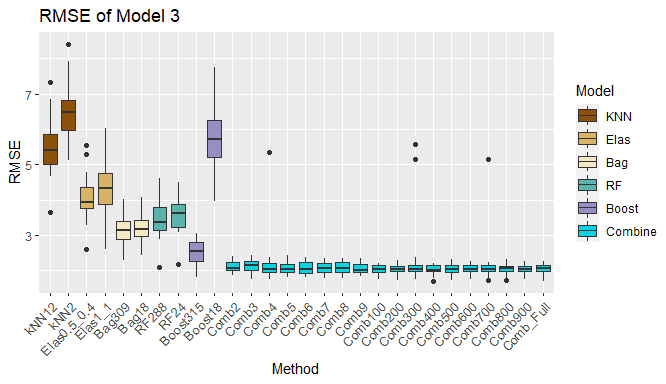}
\includegraphics[width=0.49\linewidth]{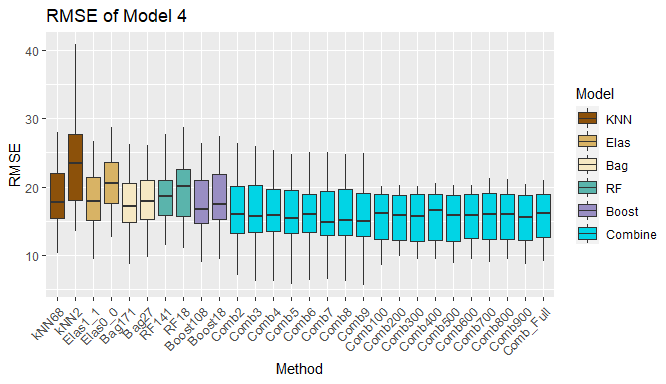}
\includegraphics[width=0.49\linewidth]{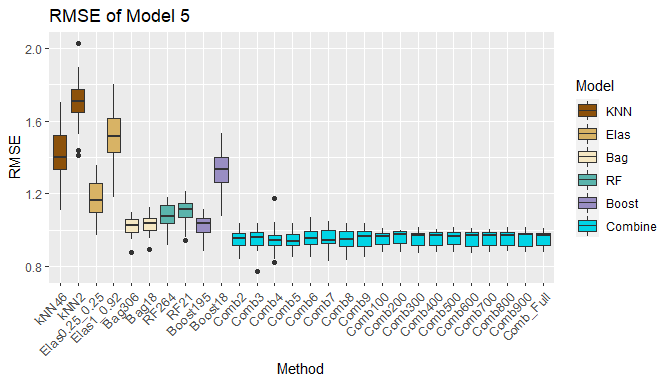}
\caption{Boxplots of average RMSEs computed on simulated datasets. From left to right, the first ten boxplots are the best and the worst performance of \textcolor{cyan}{\bf kNN}, \textcolor{cyan}{\bf Elas}, \textcolor{cyan}{\bf Bag}, \textcolor{cyan}{\bf RF} and \textcolor{cyan}{\bf Boost} machines respectively. The last eighteen boxplots represent the performances of the aggregation methods \textcolor{cyan}{\bf Comb$m$} with $m=2,3,...,9,100,200,...,900$ and \textcolor{cyan}{\bf Comb\_Full} respectively. The full aggregation performs well on $1000$ dimensional predicted features (very highly correlated). Moreover, the performances of the aggregation scheme on much lower dimensional subspaces are almost preserved compared to the full aggregation with slightly larger variances.}
\label{fig:boxErrorSimulated}
\end{figure}

\begin{remark}
Note that in all simulations, smoothing parameter $h$ is estimated using gradient descent algorithm discussed in \cite{has2021}. In all cases, the same learning rate is used, that is why on some datasets, the algorithm struggles around the optimal values of parameter, leading to slower computational times (Model~\ref{mod:4} and Model~\ref{mod:5} of Figure~\ref{fig:boxTimeSimulated}). In real situation, this can be improved by choosing more suitable values of parameter in the optimization method for any given datasets.
\end{remark}

\FloatBarrier
\subsection{Real datasets}
We consider in this section two public datasets (available and easily accessible on the internet) and two private energy datasets. The first dataset called {\bf Abalone} (available at \cite{abaloneData}) contains $4177$ rows and $9$ columns of measurements of abalones observed in Tasmania, Australia. We are interested in predicting the age of each abalone through the number of rings ({\it Rings}) using its physical characteristics such as {\it gender, size, weight},  etc. The second dataset, named {\bf Boston}, is available in \texttt{MASS} library of \texttt{R} software (see \cite{massR}), comprises of $14$ columns corresponding to median house prices ({\it medv}) and other variables of 506 suburbs in Boston such as per capita crime rate ({\it crim}), average number of rooms per dwelling ({\it rm}), pupil-teacher ratio by town ({\it ptratio}), nitrogen oxides concentration ({\it ox}), etc. Then, the goal is to predict the median house prices of those suburbs using all quantitative characteristics. 

The third dataset ({\bf Air}) considered in this section is a private dataset containing six columns corresponding to \textit{Air temperature, Input Pressure, Output Pressure, Flow, Water Temperature} and \textit{Power Consumption}, along with $2\ 026$ rows of hourly observations of these measurements of an air compressor machine provided by \cite{CHM}. The goal is to predict the power consumption of this machine using the five remaining explanatory variables. The last dataset ({\bf Turbine}) is provided by the wind energy company Ma$\ddot{\text{\i}}$a Eolis. It contains $8\ 721$ observations of seven variables representing 10-minute measurements of \textit{Electrical power, Wind speed, Wind direction, Temperature, Variance of wind speed} and \textit{Variance of wind direction} measured from a wind turbine of the company (see \cite{wind}). In this case, we aim at predicting the electrical power produced by the turbine using the remaining six measurements as explanatory variables. 

The performances obtained from $30$ independent runs, computed using the same computer mentioned in the previous section, are provided in Table~{\ref{tab:real}} below. We observe that the performances of the aggregation methods approach, and sometimes outperform the best estimator on all datasets. Moreover, all the aggregation methods perform equally well in each case regardless of the size of projected dimension. In addition, the performances (the best and the worst cases) of all machines and the aggregation methods are summarized in boxplots of Figure~{\ref{fig:boxErrorReal}} below. Finally, Figure~{\ref{fig:boxTimeReal}} illustrates time efficiency of the proposed methods.

\begin{sidewaystable}[!ph]
\scriptsize
\centering  
\hspace{0.5em}
\setlength{\tabcolsep}{3.5pt}
\def\arraystretch{1.2}
\begin{tabular}{ c | c c c c c | c c c c c c c c c}  
\hline\hline                       
\multirow{2}{*}{\textcolor{cyan}{\bf Model}} & \multicolumn{5}{c|}{\bf Basic machines} & \multicolumn{9}{c}{{\bf Aggregation method} \textcolor{cyan}{\bf Comb$m$}}\\ [0.5ex] \cline{2-15}
 & \textcolor{cyan}{\bf $k$NN} &\textcolor{cyan}{\bf Elas} &\textcolor{cyan}{\bf Bag} & \textcolor{cyan}{\bf RF} & \textcolor{cyan}{\bf Boost} & \textcolor{cyan}{\bf 100/2} & \textcolor{cyan}{\bf 200/3} & \textcolor{cyan}{\bf 300/4} & \textcolor{cyan}{\bf 400/5}  & \textcolor{cyan}{\bf 500/6} & \textcolor{cyan}{\bf 600/7} & \textcolor{cyan}{\bf 700/8} & \textcolor{cyan}{\bf 800/9} & \textcolor{cyan}{\bf 900/Comb\_Full}\\ [0.5ex]   
\hline 
\multirow{4}{*}{\bf{Abalone}} & \multirow{2}{*}{\raisebox{-2.5ex}{\bf 2.052}} & \multirow{2}{*}{\raisebox{-2.5ex}{2.092}} & \multirow{2}{*}{\raisebox{-2.5ex}{2.174}} & \multirow{2}{*}{\raisebox{-2.5ex}{  2.213}} & $\multirow{2}{*}{\raisebox{-2.5ex}{2.106}}$ & {\bf 2.135} & 2.105 & 2.114 & 2.113 & 2.113 & 2.115 & 2.112 & 2.114 & 2.113\\ 
& & & & & & (0.051) & (0.046) & (0.051) & (0.047) & (0.048) & (0.045) & (0.049) & (0.044) & (0.047)\\ \cline{7-15}
& (0.061) & (0.055) & (0.060) & (0.052) & (0.055) & 2.198 & 2.165 & 2.143 & 2.144 & 2.156 & 2.138 & 2.149 & 2.152 & 2.114\\ 
& & & & & & (0.155) & (0.095) & (0.066) & (0.061) & (0.067) & (0.067) & (0.078) & (0.063) & (0.044)\\
 \hline 
 \multirow{4}{*}{\bf{Boston}} & \multirow{2}{*}{\raisebox{-2.5ex}{6.855}} & \multirow{2}{*}{\raisebox{-2.5ex}{5.039}} & \multirow{2}{*}{\raisebox{-2.5ex}{4.410}} & \multirow{2}{*}{\raisebox{-2.5ex}{\bf 3.574}} & $\multirow{2}{*}{\raisebox{-2.5ex}{3.811}}$ & {3.048} & {\bf 3.039} & 3.073 & 3.041 & 3.055 & 3.043 & 3.049 & 3.049 & 3.051\\
& & & & & & (0.351) & (0.348) & (0.378) & (0.376) & (0.373) & (0.369) & (0.372) & (0.352) & (0.383)\\ \cline{7-15}
&(0.547) & (0.576) & (0.468) & (0.402) & (0.437) & 4.033 & 3.431 & 3.436 & 3.459 & 3.227 & 3.344 & 3.198 & 3.293 & 3.044\\
& & & & & & (1.099) & (0.724) & (0.672) & (0.596) & (0.737) & (0.631) & (0.555) & (0.679) &(0.362) \\
\hline  
\multirow{4}{*}{\bf{Air}} & \multirow{2}{*}{\raisebox{-2.5ex}{291.435}} & \multirow{2}{*}{\raisebox{-2.5ex}{177.581}} & \multirow{2}{*}{\raisebox{-2.5ex}{341.514}} & \multirow{2}{*}{\raisebox{-2.5ex}{210.910}} & $\multirow{2}{*}{\raisebox{-2.5ex}{\bf 153.538}}$ & 136.424 & 136.535 & 136.532 & 136.487 & 135.961 & 136.424 & 136.108 & 136.509 & {\bf 136.075}\\
& & & & & & (3.178) & (4.276) & (4.535) & (4.122) & (3.704) & (4.383) & (4.580) & (4.237) & (4.507)\\
\cline{7-15}
 & (9.084) & (4.763) & (16.110) & (15.899) & (5.868) & 169.592 & 151.757 & 148.344 & 146.905 & 144.371 & 143.118 & 142.619 & 143.028 & 136.828\\
& & & & & & (20.127) & (9.602) & (5.556) & (7.005) & (6.294) & (4.599) & (4.572) & (5.743) & (3.616)\\
 \hline  
 \multirow{2}{*}{\bf{Turbine}} & \multirow{2}{*}{\raisebox{-2.5ex}{39.348}} & \multirow{2}{*}{\raisebox{-2.5ex}{67.978}} & \multirow{2}{*}{\raisebox{-2.5ex}{68.110}} & \multirow{2}{*}{\raisebox{-2.5ex}{\bf 35.932}} & $\multirow{2}{*}{\raisebox{2ex}{39.850}}$ & 36.968 & 36.671 & 36.694 & 36.602 & 36.675 & 36.568 & 36.643 & 36.635 & 36.622\\
& & & & &  & (1.127) & (1.146) & (1.099) & (1.148) & (1.184) & (1.092) & (1.034) & (1.123) & (1.125)\\ \cline{7-15}
 & (1.119) & (2.505) & (1.498) & (1.038) & (0.976) & 38.916 & 37.843 & 37.390 & 37.183 & 36.970 & 36.542 & 36.673 & 36.490 & 36.465\\
& & & & &  & (2.363) & (1.201) & (1.228) & (1.244) & (1.035) & (0.745) & (0.759) & (0.880) & (1.117)\\
   \hline  
  \hline  
\end{tabular}  
\caption{Average RMSEs of real-life datasets.}
\label{tab:real}
\end{sidewaystable}

\begin{figure}[!ph]
\centering
\includegraphics[width=0.49\linewidth]{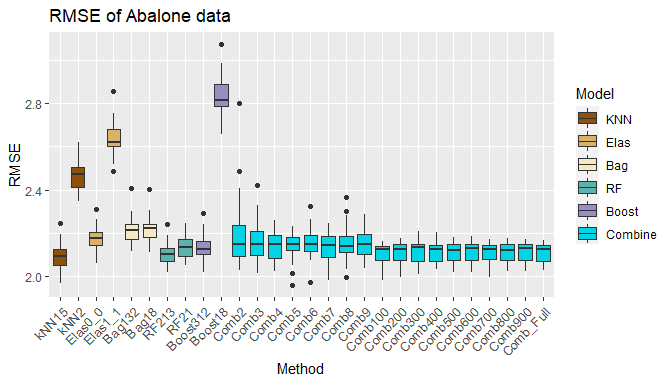}
\includegraphics[width=0.49\linewidth]{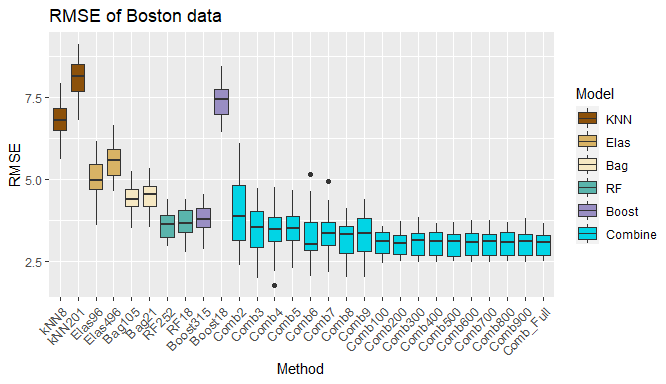}
\includegraphics[width=0.49\linewidth]{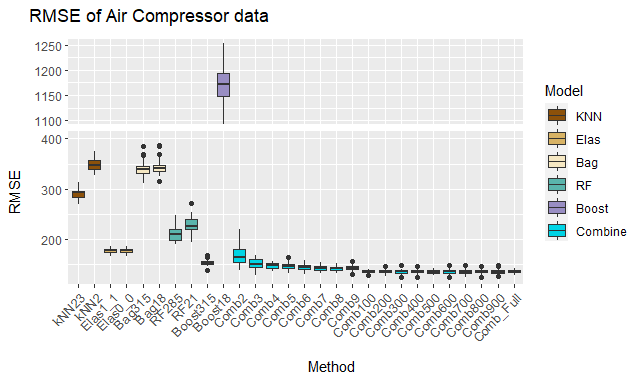}
\includegraphics[width=0.49\linewidth]{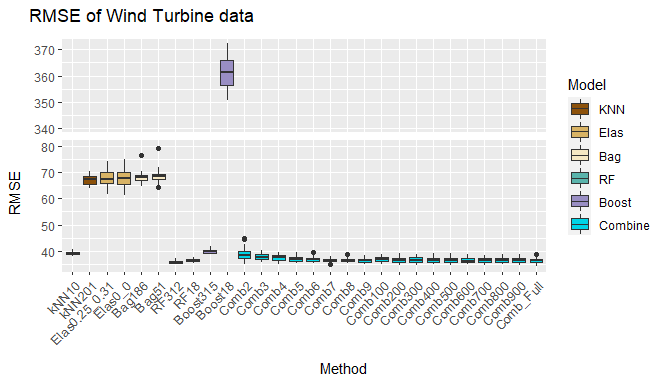}
\caption{Boxplots of average RMSEs computed on real-life datasets.}
\label{fig:boxErrorReal}
\end{figure}
\begin{figure}[!ph]
\centering
\includegraphics[width=0.49\linewidth]{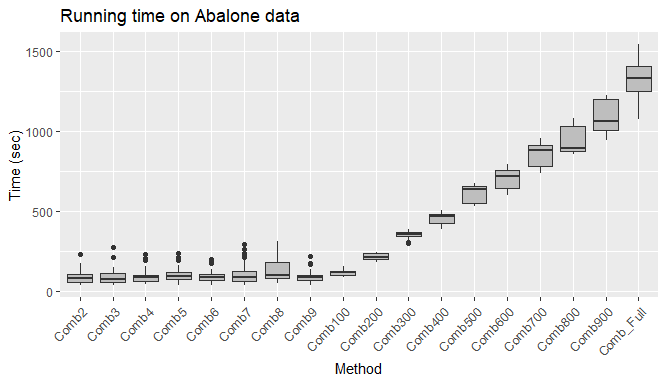}
\includegraphics[width=0.49\linewidth]{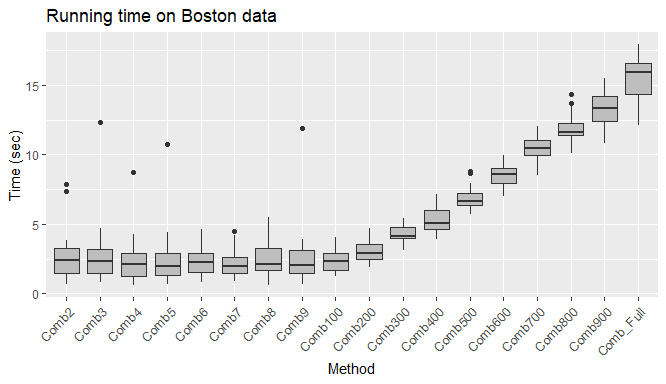}
\includegraphics[width=0.49\linewidth]{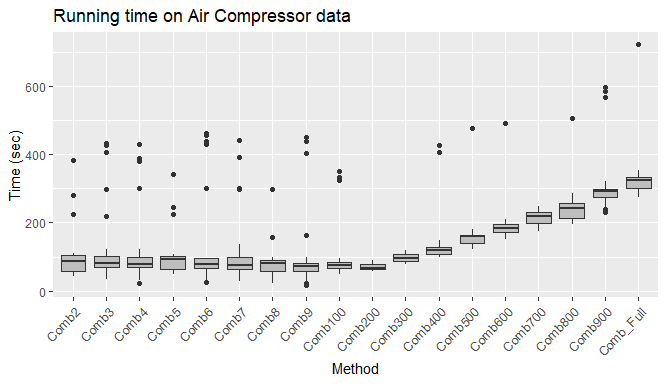}
\includegraphics[width=0.49\linewidth]{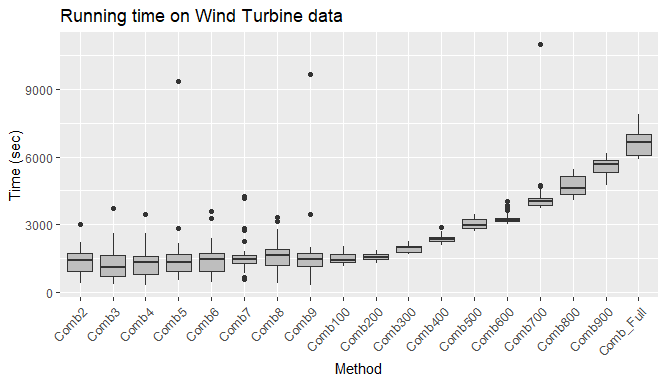}
\caption{Running times of the combining methods on real-life datasets.}
\label{fig:boxTimeReal}
\end{figure}

\section{Conclusion}
\label{sec:chp4Conclusion}
This chapter fills the gap by studying high-dimensional case of consensual aggregation for regression. The aggregation scheme is composed of two steps: high-dimensional features of predictions are first random projected into a smaller space using Johnson-Lindenstrauss method, then the exponential kernel-based aggregation method is implemented on the projected features. First, we theoretically show that the performance of the projected and full aggregation methods are close, with high probability. Then, we numerically illustrate that the full aggregation method upholds its performance on very large redundant features given by different types of predictors. Together, this indicates the robustness of the method in a sense that, one can plainly construct several types of predictive models with different values of parameters in parallel, then flexibly aggregate them directly without any model validation step. All these results are confirmed through several numerical experiments carried out on different types of simulated and real datasets. On top of that, in term of computational speed, the proposed method is often much faster (from 3 to 20 times) compared to the full aggregation method according the optimization process (learning rate, for instance). 
\section{Proofs}
\label{sec:ch4proof}
\subsection{Proof of proposition \ref{prop:JL1}}
\label{subsec:proofProp1}
Under the assumption of the proposition, using the results of (\ref{eq:JL1}), (\ref{eq:JL2}) and (\ref{eq:inequality}), the union bound probability implies for any $\delta\in (0,1)$:
\begin{align*}
&\quad\ \mathbb{P}\Big(\exists z_j\in S_n:\Big|\frac{\|\tilde{z_0}-\tilde{z_j}\|^2}{\|z_0-z_j\|^2}-1\Big|>\delta\Big)\\
&=\mathbb{P}\Big(\exists z_j\in S_n:\frac{\|\tilde{z_0}-\tilde{z_j}\|^2}{\|z_0-z_j\|^2}-1>\delta\Big) +\mathbb{P}\Big(\exists z_j\in S_n:\frac{\|\tilde{z_0}-\tilde{z_j}\|^2}{\|z_0-z_j\|^2}-1<-\delta\Big)\\
&\leq \sum_{j=1}^n\mathbb{P}\Big(\frac{\|\tilde{z_0}-\tilde{z_j}\|^2}{\|z_0-z_j\|^2}-1>\delta\Big) + \sum_{j=1}^n\mathbb{P}\Big(\frac{\|\tilde{z_0}-\tilde{z_j}\|^2}{\|z_0-z_j\|^2}-1<-\delta\Big) \\
&\leq \sum_{j=1}^ne^{m[-\delta+\ln(1+\delta)]/2}+\sum_{j=1}^ne^{m[\delta+\ln(1-\delta)]/2}\\
&\leq ne^{-m(\delta^2/2-\delta^3/3)/2}+ne^{-m(\delta^2/2+\delta^3/3)/2}\\
&\leq 2ne^{-m(\delta^2/2-\delta^3/3)/2}.
\end{align*}
We conclude the proof using the complementary probability,
$$\mathbb{P}\Big(\Big|\frac{\|\tilde{z_0}-\tilde{z_j}\|^2}{\|z_0-z_j\|^2}-1\Big|\leq\delta,\forall z_j\in S_n\Big)\geq 1-2ne^{-m(\delta^2/2-\delta^3/3)/2}.$$
\QED

\subsection{Proof of Theorem \ref{prop:question1}}
For the sake of readability, for any $j=1,2,...,n$, let
\begin{itemize}
\item $K_h^j=K_h(\|{{\bf r}}(X)-{{\bf r}}(X_j)\|)$.
\item  $\tilde{K}_h^j=K_h(\|\tilde{{\bf r}}(X)-\tilde{{\bf r}}(X_j)\|)$.
\end{itemize}
For any $x\in\mathbb{R}^d$ and for any $h>0$,
\begin{align*}
|g_n(\textbf{r}(X))-g_n(\tilde{\textbf{r}}(X))|&=\Big|\frac{\sum_{i=1}^nY_iK_h^i}{\sum_{j=1}^nK_h^j}-\frac{\sum_{i=1}^nY_i\tilde{K}_h^i}{\sum_{j=1}^n\tilde{K}_h^j}\Big|\\
&=\Big|\frac{\sum_{i=1}^nY_iK_h^i}{\sum_{j=1}^nK_h^j}-\frac{\sum_{i=1}^nY_i\tilde{K}_h^i}{\sum_{j=1}^nK_h^j}+\frac{\sum_{i=1}^nY_i\tilde{K}_h^i}{\sum_{j=1}^nK_h^j}-\frac{\sum_{i=1}^nY_i\tilde{K}_h^i}{\sum_{j=1}^n\tilde{K}_h^j}\Big|\\
&\leq R_0\frac{\sum_{i=1}^n|K_h^i-\tilde{K}_h^i|}{\sum_{j=1}^nK_h^j}+R_0\Big[\sum_{j=1}^n\tilde{K}_h^j\Big]\frac{|\sum_{i=1}^n\tilde{K}_h^i-\sum_{i=1}^nK_h^i|}{\Big[\sum_{j=1}^nK_h^i\Big]\Big[\sum_{j=1}^n\tilde{K}_h^j\Big]}\\
&\leq R_0\frac{\sum_{i=1}^n|K_h^i-\tilde{K}_h^i|}{\sum_{j=1}K_h^j}+R_0\frac{\sum_{i=1}^n|\tilde{K}_h^i-K_h^i|}{\sum_{j=1}^nK_h^j}\\
&= 2R_0\frac{\sum_{i=1}^n|K_h^i-\tilde{K}_h^i|}{\sum_{j=1}^nK_h}\\
&= 2R_0\frac{\sum_{i=1}^nK_h^i|1-\tilde{K}_h^i/K_h^i|}{\sum_{j=1}^nK_h^j}\\
&\leq 2R_0\max_{1\leq i\leq n}\Big|1-\frac{\tilde{K}_h^i}{K_h^i}\Big|.
\end{align*}

Therefore, for any $\varepsilon>0$, one has:

\begin{align*}
&\quad\ \mathbb{P}\Big(|g_n(\textbf{r}(X))-g_n(\tilde{\textbf{r}}(X))|>\varepsilon\Big)\\
&\leq\mathbb{P}\Big(2R_0\max_{1\leq i\leq n}\Big|1-\frac{K_h(\|\tilde{{\bf r}}(X)-\tilde{{\bf r}}(X_i)\|)}{K_h(\|{\bf r}(X)-{\bf r}(X_i)\|)}\Big|>\varepsilon\Big)\\
&=1-\mathbb{P}\Big(2R_0\max_{1\leq i\leq n}\Big|1-\frac{K_h(\|\tilde{{\bf r}}(X)-\tilde{{\bf r}}(X_i)\|)}{K_h(\|{\bf r}(X)-{\bf r}(X_i)\|)}\Big|\leq\varepsilon\Big).
\end{align*}
One can compute the last probability using independency of $(X_i)_{i=1}^n$ and Fubini's theorem as follow

\begin{align*}
&\quad\ \mathbb{P}\Big(2R_0\max_{1\leq i\leq n}\Big|1-\frac{K_h(\|\tilde{{\bf r}}(X)-\tilde{{\bf r}}(X_i)\|)}{K_h(\|{\bf r}(X)-{\bf r}(X_i)\|)}\Big|\leq\varepsilon\Big)\\
&=\int_{\mathbb{R}^{M}}\int_{\mathbb{R}^{M\times m}}\mathbb{P}_{(X_i)_{i=1}^n}\Big(2R_0\max_{1\leq i\leq n}\Big|1-\frac{K_h(\|({\bf r}(x)-{\bf r}(X_i))G\|)}{K_h(\|{\bf r}(x)-{\bf r}(X_i)\|)}\Big|\leq\varepsilon\Big)\mathbb{P}_{\mathcal{G}}(G)\mu(dx)\\
&=\int_{\mathbb{R}^{M}}\int_{\mathbb{R}^{M\times m}}\Big[\mathbb{P}_{X_1}\Big(2R_0\Big|1-\frac{K_h(\|({\bf r}(x)-{\bf r}(X_1))G\|)}{K_h(\|{\bf r}(x)-{\bf r}(X_1)\|)}\Big|\leq\varepsilon\Big)\Big]^n\mathbb{P}_{\mathcal{G}}(G)\mu(dx)\\
&=\int_{\mathbb{R}^{M}}\int_{\mathbb{R}^M}\Big[\mathbb{P}_{\mathcal{G}}\Big(2R_0\Big|1-\frac{K_h(\|({\bf r}(x)-{\bf r}(v))G\|)}{K_h(\|{\bf r}(x)-{\bf r}(v)\|)}\Big|\leq\varepsilon\Big)\Big]^n\mu(dv)\mu(dx)\\
&\geq \Big[\int_{\mathbb{R}^{M}}\int_{\mathbb{R}^M}\mathbb{P}_{\mathcal{G}}\Big(2R_0\Big|1-\frac{K_h(\|({\bf r}(x)-{\bf r}(v))G\|)}{K_h(\|{\bf r}(x)-{\bf r}(v)\|)}\Big|\leq\varepsilon\Big)\mu(dv)\mu(dx)\Big]^n.
\end{align*}
The last bound of the above inequality is obtained by Jensen's inequality. Next, for any $x,v\in\mathbb{R}^d$, given all the basic machines $(r_k)_{k=1}^M$, Johnson-Lindenstrauss Lemma implies that for any $\delta_0\in(0,1)$, with probability at least $1-2e^{-m(\delta_0^2/2-\delta_0^3/3)/2}$, one has:
\begin{align*}
\Big|\frac{\|\tilde{{\bf r}}(x)-\tilde{{\bf r}}(v)\|^2}{\|{\bf r}(x)-{\bf r}(v)\|^2}-1\Big|&\leq\delta_0\\
\Leftrightarrow (1-\delta_0)\|{\bf r}(x)-{\bf r}(v\|^2\leq\|\tilde{{\bf r}}(x)-\tilde{{\bf r}}(v)\|^2&\leq(1+\delta_0)\|{\bf r}(x)-{\bf r}(X_j)\|^2\\
\Leftrightarrow (1-\delta_0)^{\alpha/2}\|{\bf r}(x)-{\bf r}(v)\|^{\alpha}\leq\|\tilde{{\bf r}}(x)-\tilde{{\bf r}}(v)\|^{\alpha}&\leq(1+\delta_0)^{\alpha/2}\|{\bf r}(x)-{\bf r}(v)\|^{\alpha}.
\end{align*}
Thus for any $x,v\in\mathbb{R}^d$, with probability at least $1-2e^{-m(\delta_0^2/2-\delta_0^3/3)/2}$ such that
\begin{align*}
\Big|\frac{K_h(\|\tilde{\bf r}(x)-\tilde{\bf r}(v)\|)}{K_h(\|{\bf r}(x)-{\bf r}(v))}-1\Big|
&\leq \exp\Big[-(\|(\tilde{{\bf r}}(x)-\tilde{{\bf r}}(v))/h\|^{\alpha}-\|({\bf r}(x)-{\bf r}(v))/h\|^{\alpha})/\sigma\Big]-1\\
&\leq \exp\Big((1-(1-\delta_0)^{\alpha/2})\|({\bf r}(x)-{\bf r}(v))/h\|^{\alpha}/\sigma\Big)-1\\
&\leq \exp\Big((1-(1-\delta_0)^{\alpha/2})(2R_0/h)^{\alpha}/\sigma\Big)-1\\
&\leq \exp\Big(\delta_0(1+\alpha/2)(2R_0/h)^{\alpha}/\sigma\Big)-1,
\end{align*}
where the last inequality above is obtained using the following inequality:
$$1-(1-\delta_0)^{\alpha}\leq \delta_0(1+\alpha),\forall \delta_0\in(0,1),\forall \alpha>0.$$
And if one take $\varepsilon=2R_0\Big(\exp\Big(\delta_0(1+\alpha/2)(2R_0/h)^{\alpha}/\sigma\Big)-1\Big)$, thus
\begin{align*}
\delta_0&=\frac{\sigma\ln(1+\varepsilon/(2R_0))}{(1+\alpha/2)(2R_0)^{\alpha}}h^{\alpha}\\
&=C_0\frac{\sigma\varepsilon h^{\alpha}}{(1+\alpha/2)(2R_0)^{1+\alpha}},
\end{align*}
where the constant $C_0\approx 1$ for small $\varepsilon>0$, and will be ignored. Therefore, for any $x,v\in\mathbb{R}^d$, and using the fact that for any $\delta_0\in(0,1):\delta_0^2/2-\delta_0^3/3\geq \delta_0^2/6$, one has
\begin{align*}
\mathbb{P}_{\mathcal{G}}\Big(2R_0\Big|1-\frac{K_h(\|({\bf r}(x)-{\bf r}(v))G\|)}{K_h(\|{\bf r}(x)-{\bf r}(v)\|)}\Big|\leq\varepsilon\Big)&\geq 1-2\exp\Big(-\frac{m(\delta_0^2/2-\delta_0^3/3)}{2}\Big)\\
&\geq 1-2\exp\Big(-\frac{m\delta_0^2}{12}\Big)\\
&\geq 1-2\exp\Big[-\frac{m(\sigma h^{\alpha}\varepsilon)^2}{3(2+\alpha)^2(2R_0)^{2(\alpha+1)}}\Big]\\
&= 1-2\exp\Big(-\frac{mh^{2\alpha}\varepsilon^2}{C_1}\Big),
\end{align*}
where the constant $C_1=3(2+\alpha)^2(2R_0)^{2(\alpha+1)}>0$. Therefore, one has
\begin{align*}
&\mathbb{P}\Big(2R_0\max_{1\leq i\leq n}\Big|1-\frac{K_h(\|\tilde{{\bf r}}(X)-\tilde{{\bf r}}(X_i)\|)}{K_h(\|{\bf r}(X)-{\bf r}(X_i)\|)}\Big|\leq\varepsilon\Big)\geq \Big[1-2\exp\Big(-\frac{mh^{2\alpha}\varepsilon^2}{C_1}\Big)\Big]^n.
\end{align*}
And this implies
\begin{align*}
&\quad\ \mathbb{P}\Big(|g_n(\textbf{r}(X))-g_n(\tilde{\textbf{r}}(X))|>\varepsilon\Big)\\
&\leq\mathbb{P}\Big(2R_0\max_{1\leq i\leq n}\Big|1-\frac{K_h(\|\tilde{{\bf r}}(X)-\tilde{{\bf r}}(X_i)\|)}{K_h(\|{\bf r}(X)-{\bf r}(X_i)\|)}\Big|>\varepsilon\Big)\\
&\leq 1-\Big[1-2\exp\Big(-\frac{mh^{2\alpha}\varepsilon^2}{3R_1^{2}}\Big)\Big]^n.
\end{align*}
Thus, for any $\delta\in (0,1)$,
\begin{align*}
1-\Big[1-2\exp\Big(-\frac{mh^{2\alpha}\varepsilon^2}{3R_1^{2}}\Big)\Big]^n&\leq\delta\\
\Leftrightarrow m&\geq C_1\frac{\log[2/(1-\sqrt[n]{1-\delta})]}{h^{2\alpha}\varepsilon^2}.
\end{align*}
Moreover, for any large $n$, one has $(1-\sqrt[n]{1-\delta})\approx -\log(1-\delta)/n$, which implies that the lower bound of $m$ is approximately $$C_1\frac{\log[-2n/\log(1-\delta)]}{h^{2\alpha}\varepsilon^2}.$$
Moreover, for small $\delta$, the order of this bound is roughly
$$O\Big(\frac{\log(2n/\delta)}{h^{2\alpha}\varepsilon^2}\Big).$$
\QED

\end{document}